\documentclass[letterpaper]{article} % DO NOT CHANGE THIS
\usepackage{aaai24}
\usepackage{booktabs}

\usepackage{float}
\usepackage{adjustbox}

\usepackage{times}  % DO NOT CHANGE THIS
\usepackage{helvet}  % DO NOT CHANGE THIS
\usepackage{courier}  % DO NOT CHANGE THIS
\usepackage[hyphens]{url}  % DO NOT CHANGE THIS
\usepackage{graphicx} % DO NOT CHANGE THIS
\usepackage[numbers]{natbib}  % DO NOT CHANGE THIS AND DO NOT ADD ANY OPTIONS TO IT

\usepackage{caption} % DO NOT CHANGE THIS AND DO NOT ADD ANY OPTIONS TO IT
\usepackage{algorithm}
\usepackage{algorithmic}
\usepackage{xcolor}
\usepackage{color}
\usepackage{url}

\usepackage{multirow}
\usepackage{makecell}
\usepackage{amssymb}
\usepackage{pifont}

\usepackage{newfloat}
\usepackage{listings}
\DeclareCaptionStyle{ruled}{labelfont=normalfont,labelsep=colon,strut=off} % DO NOT CHANGE THIS
\floatstyle{ruled}
\newfloat{listing}{tb}{lst}{}
\floatname{listing}{Listing}
\usepackage{hyperref} %-- This package is specifically forbidden
\usepackage{verbatim}
\usepackage{markdown}

\title{SuperCLUE-Fin: Graded Fine-Grained Analysis of Chinese LLMs on Diverse Financial Tasks and Applications}
\author{
    Liang Xu, \ %\textsuperscript{\rm 1},
    Lei Zhu, \ %\textsuperscript{\rm 1},
    Yaotong Wu, \ %\textsuperscript{\rm 1},
    Hang Xue\ %\textsuperscript{\rm 1}
}
\affiliations{
    SuperCLUE team
    contact@superclue.ai %, \\
}

\usepackage{bibentry}
\begin{document}
\pagestyle{plain}

\maketitle

\begin{abstract}

The SuperCLUE-Fin (SC-Fin) benchmark is a pioneering evaluation framework tailored for Chinese-native financial large language models (FLMs). It assesses FLMs across six financial application domains and twenty-five specialized tasks, encompassing theoretical knowledge and practical applications such as compliance, risk management, and investment analysis. Using multi-turn, open-ended conversations that mimic real-life scenarios, SC-Fin measures models on a range of criteria, including accurate financial understanding, logical reasoning, clarity, computational efficiency, business acumen, risk perception, and compliance with Chinese regulations.

In a rigorous evaluation involving over a thousand questions, SC-Fin identifies a performance hierarchy where domestic models like GLM-4 and MoonShot-v1-128k outperform others with an A-grade, highlighting the potential for further development in transforming theoretical knowledge into pragmatic financial solutions. This benchmark serves as a critical tool for refining FLMs in the Chinese context, directing improvements in financial knowledge databases, standardizing financial interpretations, and promoting models that prioritize compliance, risk management, and secure practices.

We create a contextually relevant and comprehensive benchmark that drives the development of AI in the Chinese financial sector. SC-Fin facilitates the advancement and responsible deployment of FLMs, offering valuable insights for enhancing model performance and usability for both individual and institutional users in the Chinese market..~\footnote{Our benchmark can be found at \url{https://www.CLUEbenchmarks.com}}.

\end{abstract}

\section{Introduction}

Science and technology finance as well as digital finance are the development trend of The Times when science and technology penetrate into the financial field. The research and development of large financial model in the rapid development of large language model is in line with the characteristics of this era and the development needs of enterprises, and provides a new technology enabling idea for the financial industry to better serve the society.

The penetration and integration of large language models into various industries is a technological development trend. As far as the financial industry is concerned, many large financial models have gradually emerged in this era. Therefore, it has become an important and necessary topic how to provide accurate quantitative evaluation criteria for the Chinese native financial large model in strict accordance with the financial supervision system and combined with the knowledge in the financial field, and timely feedback on the development of the large model.

In order to evaluate the level of development of big financial models and make suggestions for improvement, we have released the SuperCLUE-Fin (SC-Fin) Chinese native big financial Model evaluation benchmark. According to different task types, the financial model is evaluated in an all-round and multi-angle way.

\section{SuperCLUE-Fin}

\begin{figure*}[h] %[p] %h
\centering
%\raggedright
% [width=0.7\textwidth, height=0.6\textheight]
\includegraphics[width=0.9\textwidth, height=0.6\textheight]{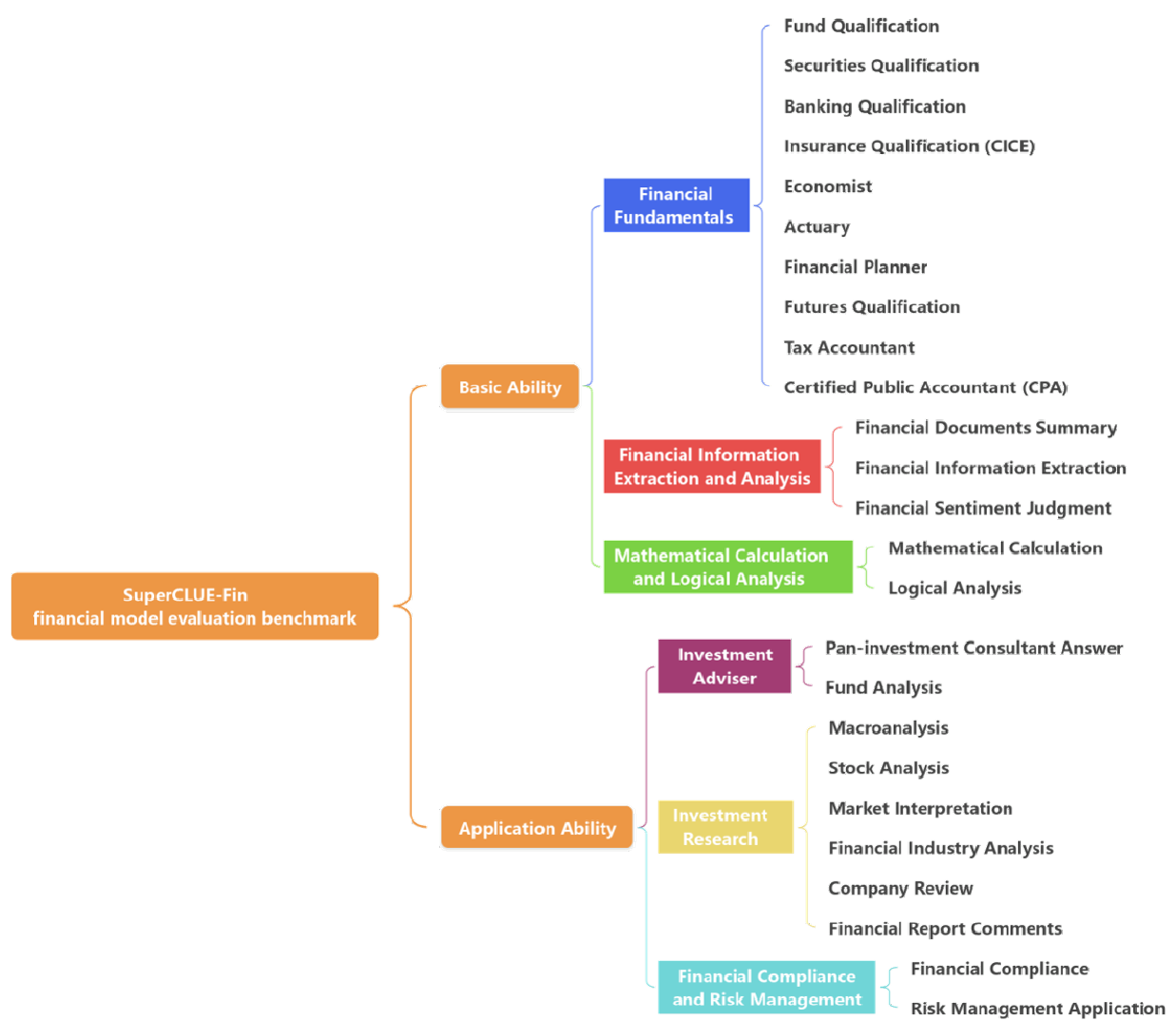}
\caption{Overview of SuperCLUE-Fin task}
\label{fig:better_performance_with_response_length_increasing}
\end{figure*}

\begin{figure*}[h] %[pt] %[pt] %p
\centering
%\raggedright
% [width=0.7\textwidth, height=0.6\textheight]
\includegraphics[width=0.8\textwidth, height=0.6\textheight]{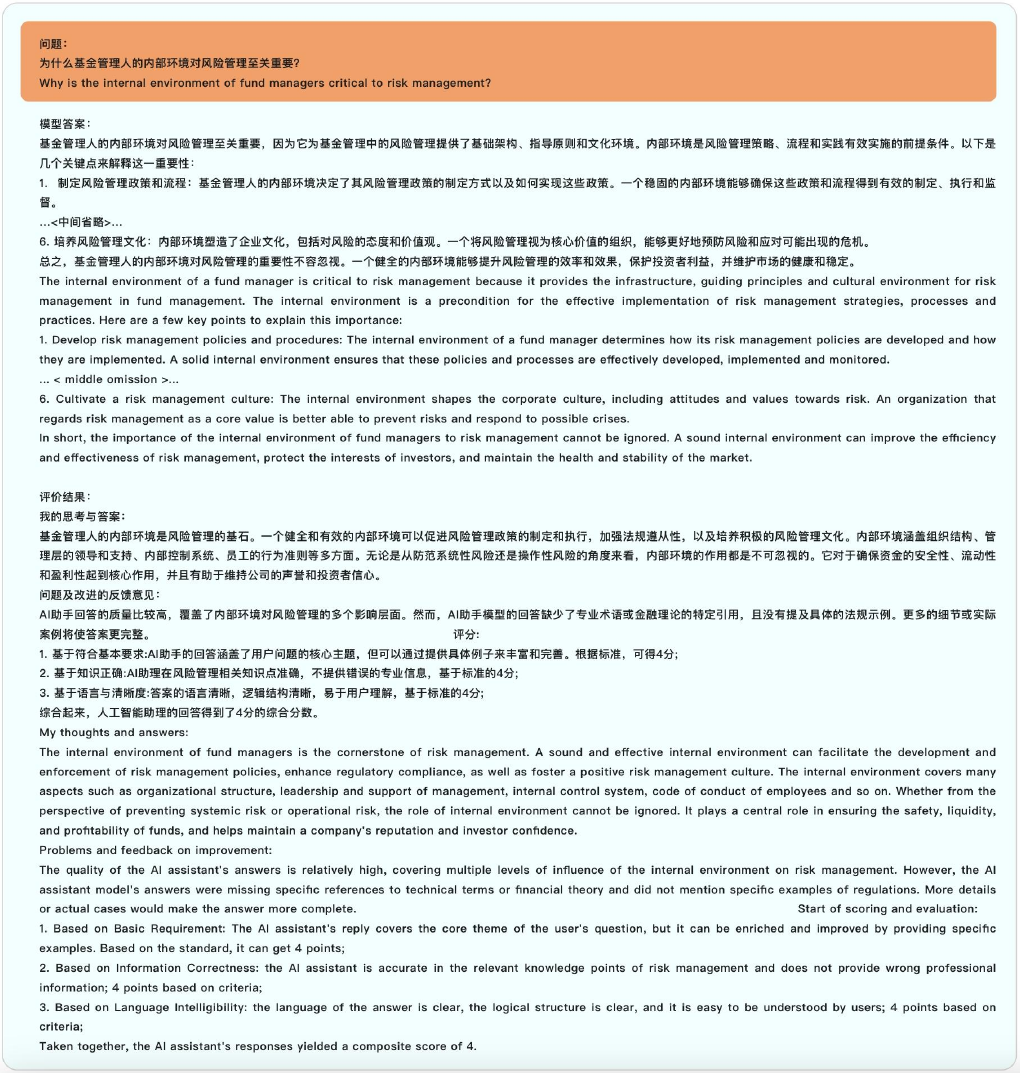}
\caption{An example of a problem in SC-Fin on QA}
\label{fig:example_math_2}
\end{figure*} 

\begin{figure*}[h]%[pt] %[pt] %p
\centering
%\raggedright
% [width=0.7\textwidth, height=0.6\textheight]
\includegraphics[width=0.8\textwidth, height=0.6\textheight]{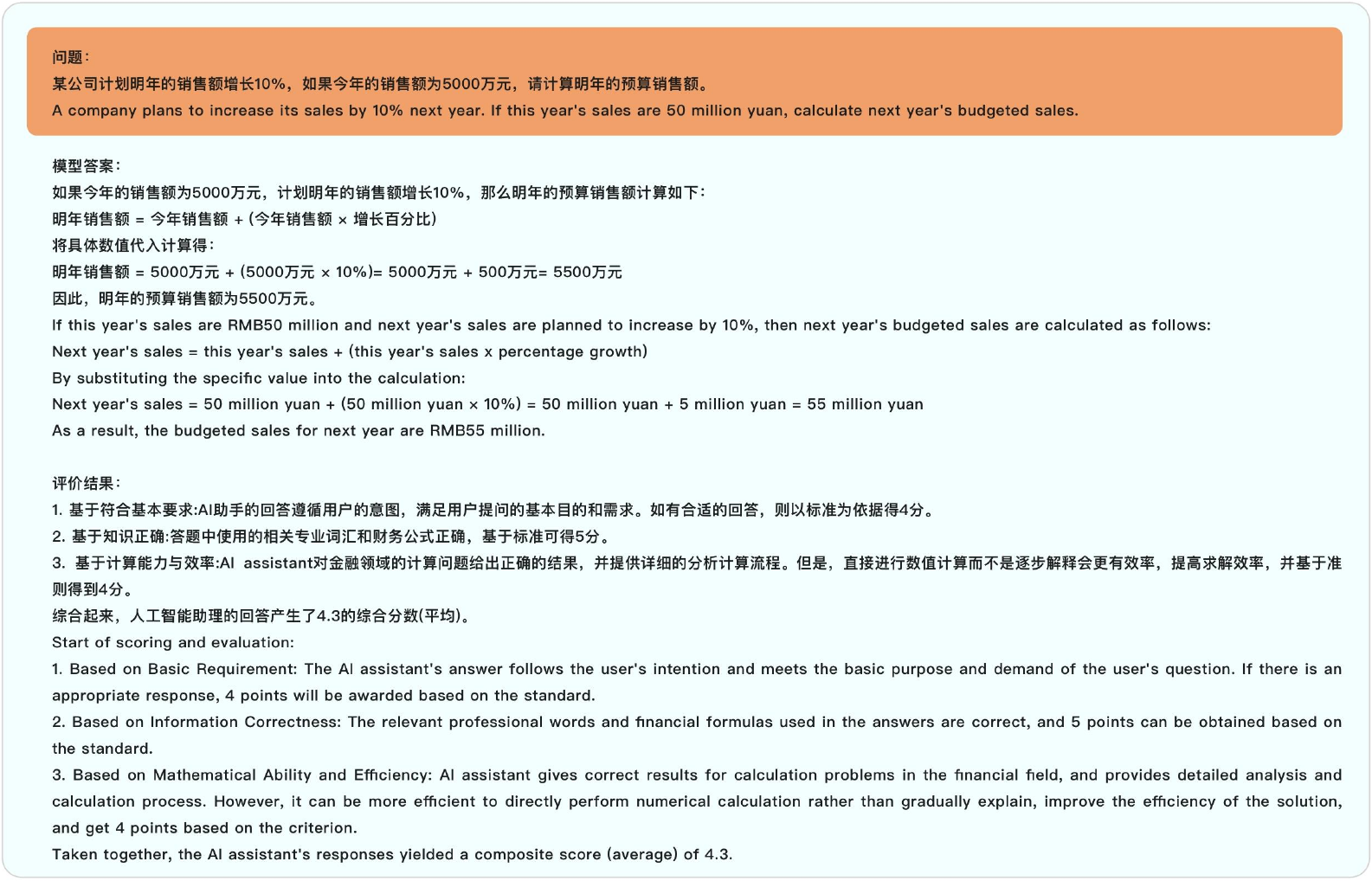}
\caption{An example of a problem in SC-Fin on the problem of calculation}
\label{fig:example_math_2}
\end{figure*} 

\begin{figure*}[h]%[pt] %[pt] %p
\centering
%\raggedright
% [width=0.7\textwidth, height=0.6\textheight]
\includegraphics[width=0.80\textwidth, height=0.65\textheight]{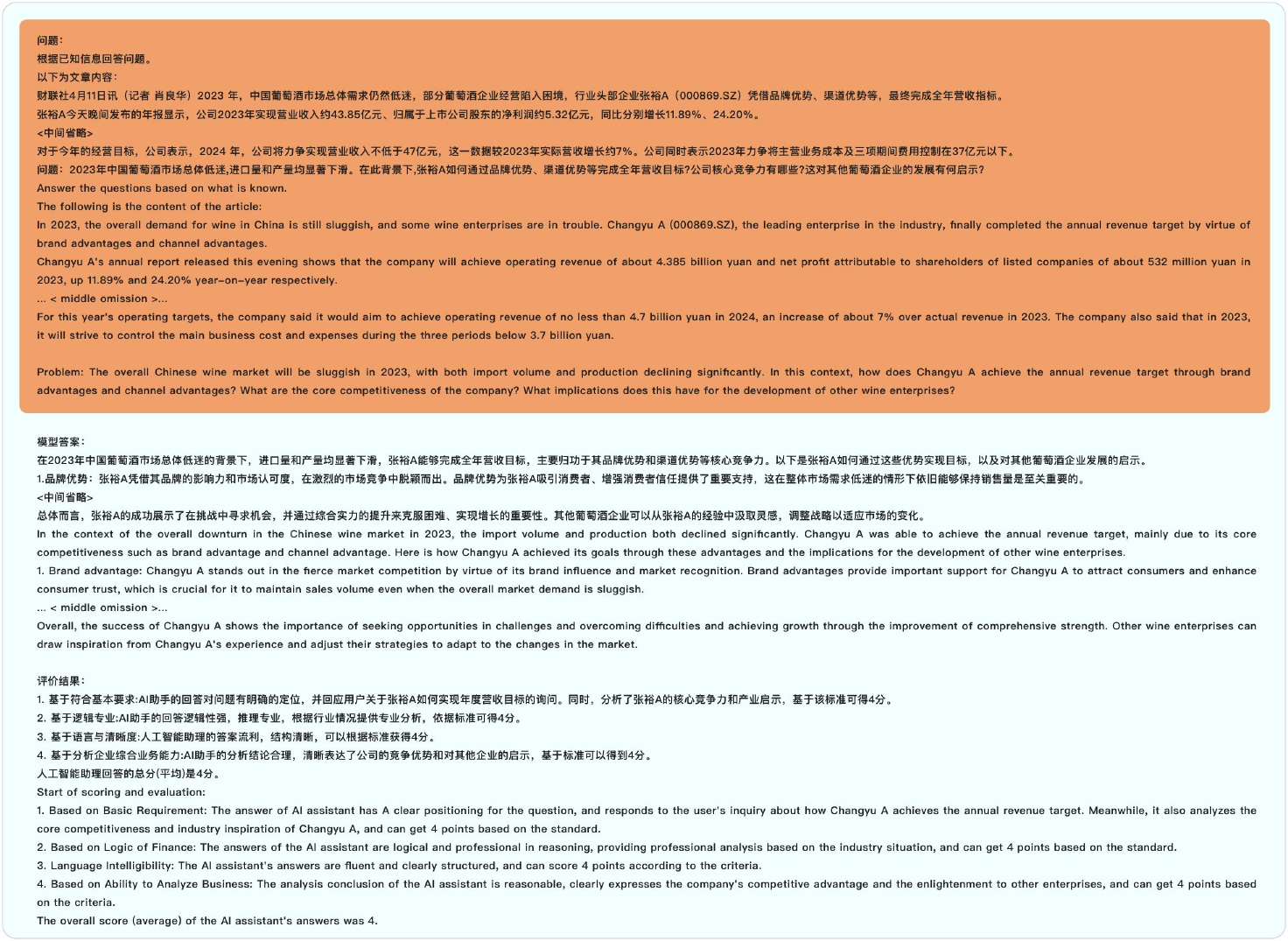}
\caption{An example of a problem in SC-Fin on professional analysis}
\label{fig:example_math_2}
\end{figure*}

\subsection{Characteristics} 

%SuperCLUE-Fin distinguishes itself with its Chinese-native financial capabilities, focusing on authentic user scenarios by utilizing entirely Chinese prompts and inputs. It encompasses a wide range of financial applications, involving six overarching categories and twenty-five detailed tasks, engaging models in diversified questioning formats, from single-turn to multi-turn interactive QA. It aims to test models’ abilities in handling everything from basic financial knowledge to complex applications in compliance, risk management, investment research, and investment consultancy.
1) Assessment of Chinese native financial ability
Based on providing Chinese world evaluation infrastructure for the era of general artificial intelligence, text input or prompt words are native Chinese; Taking full account of the development status of China's financial industry and the characteristics of China's financial regulatory system, we are committed to creating financial model evaluation indicators suitable for the Chinese context.

2) Wide range of financial application scenarios
This evaluation sets up six application scenarios related to the financial field and 25 subdivided task types, covering various practical problems from macro to micro, from abstract to specific in various industries in the financial field, aiming to investigate the comprehensive strength of the financial model in dealing with the above various tasks from an all-round and multi-perspective.

3) Various ways of investigating problems
This evaluation adopts a variety of question investigation methods, in form is divided into a single round of questions and answers, multiple rounds of questions and answers; The question types include noun explanation, calculation, brief answer, material analysis, comprehensive demonstration, etc., aiming to investigate the ability of the financial model to deal with various application problems from brief answer to complex through diversified questioning methods.

4) Open multi-round interactive QA
In order to fully restore the user experience scene, most of the questions in this evaluation adopt open multi-round interactive QA (a few are single round), and no objective questions are used as evaluation data. The purpose is to highly reflect the interactive experience between users and the large financial model, and make a more comprehensive and real investigation on the interactive QA ability of the model.

\subsection{Dataset and Task Dimensions}

The dataset contains over 1000 questions divided among numerous tasks, touching upon various financial sectors such as fund management, securities, futures, insurance, investment, wealth management, taxation, and banking. The tasks are classified into two primary dimensions:

1)Basic Capabilities: This includes tasks like Financial Fundamentals, Financial Information Extraction and Analysis, and Mathematical Calculation and Logical Analysis.

2)Applied Capabilities: This dimension focuses on tasks related to Financial Compliance and Risk Management, Investment Research, and Investment Adviser.

The content of each level task level sub-category is introduced as follows:

a. Financial Fundamentals(FF): examine the model's mastery of basic knowledge in the financial field, including funds, securities, futures, insurance, investment, financial management, taxation, banking and so on. Specific sub-category tasks include: Fund Qualification, Securities Qualification, Banking Qualification, Insurance Qualification (CICE), Economist, Actuary, Financial Planner, Futures Qualification, Tax Accountant, Certified Public Accountant (CPA).

b. Financial Information Extraction and Analysis(FIEA): examine the model's ability to interpret documents related to the financial field and its ability to judge and understand financial entities, financial intentions, financial emotions and other contents. Specific sub-tasks include: Financial Documents Summary, Financial Information Extraction, Financial Sentiment Judgment.

c. Mathematical Calculation and Logical Analysis(MCLA): examine the model's ability to identify and logically analyze actual scenarios in the financial field and its ability to skillfully use financial formulas to solve practical problems. Specific sub-tasks include:Mathematical Calculation, Logical Analysis.

d. Financial Compliance and Risk Management(FCRM): examine the model's ability to master and apply domestic financial regulatory systems and industry standards, as well as its ability to assist users in analyzing potential financial risks of business and proposing corresponding control measures. Specific sub-tasks include: Financial Compliance, Risk Management Application.

e. Investment Research(IR): investigate the model's ability to control and analyze macro economy, market trend, industry development, corporate business and stock market conditions, and assist users to make correct judgments and decisions in the investment research stage. Specific sub-tasks include: Macroanalysis, Stock Analysis, Market Interpretation, Financial Industry Analysis, Company Review, Financial Report Comments.

f. Investment Adviser(IA): inspection model as the user's investment consultant to assist users to complete the analysis and formulation of investment strategies. Specific subcategory tasks include: Pan-investment Consultant Answer, Fund Analysis.

Three categories of examples are provided as figure2-4.

\subsection{Evaluation Dimensions}

Each response is evaluated across multiple criteria, including Basic Requirement, Information Correctness, Logic of Finance, Language Intelligibility, Mathematical, Ability and Efficiency, Ability to Analyze Business, Clear Judgment, Safety Measure, Financial Security and Compliance, Risk Prediction and Control, Financial Insight. Scores are assigned on a five-point scale for each criterion, and the final model score is calculated as the average across all tasks.
\begin{figure}[h]
\centering
%\raggedright
% [width=0.7\textwidth, height=0.6\textheight]
\includegraphics[width=0.5\textwidth, height=0.2\textheight]{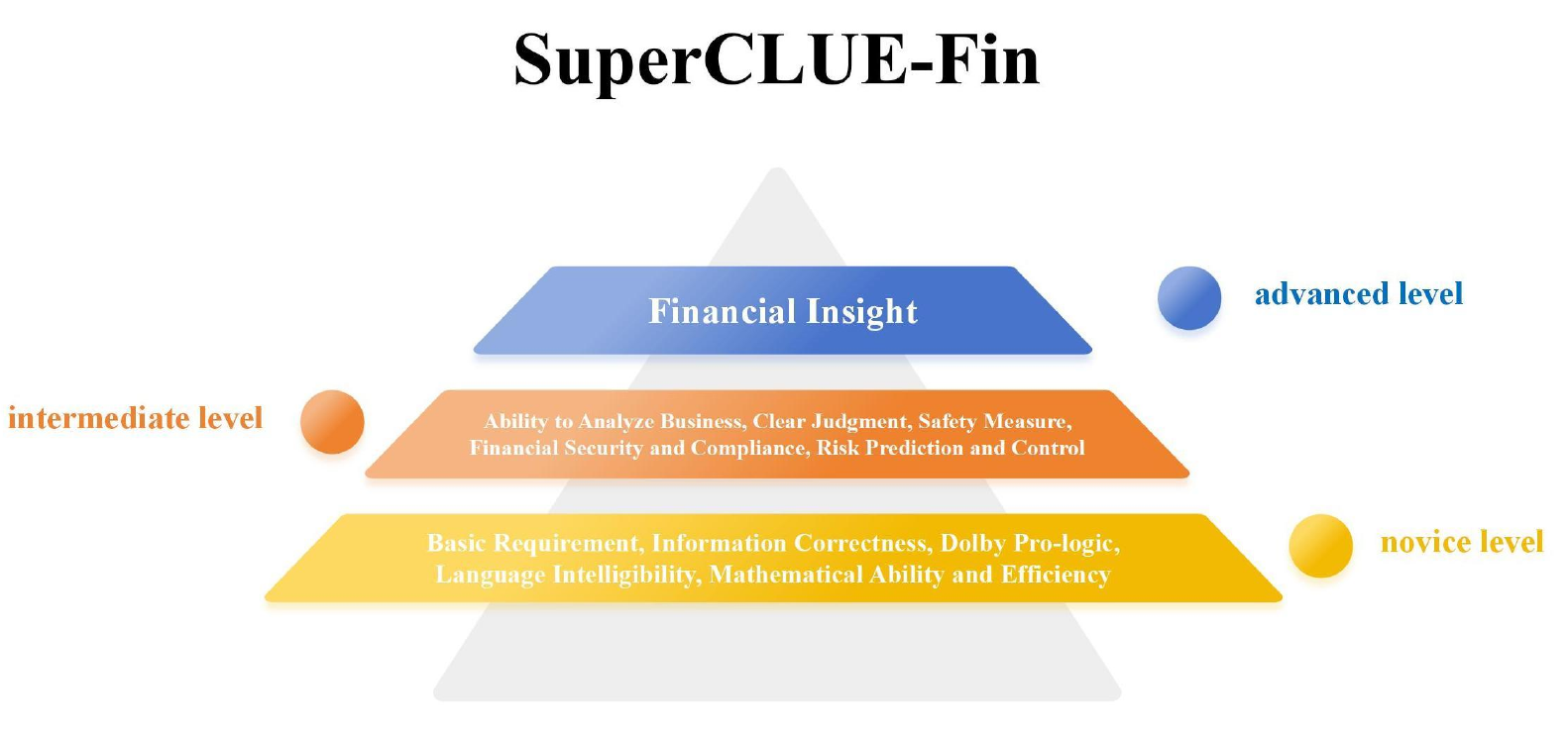}
\caption{Overview of SuperCLUE-Fin evaluation criterion}
\label{fig:better_performance_with_response_length_increasing}
\end{figure}
The definitions of the evaluation criteria are as follows:

a. Basic Requirement(BR):  The answer should follow the user's intention, meet the basic purpose and needs of the user to ask the question appropriately.

b. Information Correctness(IC): The interpretation and use of professional words involved in the answer should be completely correct, including professional terms, financial formulas, financial theories, and related knowledge in the economic field.

c. Logic of Finance(LF): The analytical logic of the answer should have a certain degree of professionalism, and demonstrate advanced thinking and expertise in financial information interpretation is needed for the perspective of solving the problem .

d.Language Intelligibility(LI): The answers are clear and easy to understand, using concise language and expressions so that users can easily understand them.

e. Mathematical Ability and Efficiency(MAE): The mathematical calculation problem in the financial field should be responded to efficiently and give correct results. At the same time, it should have a detailed analysis and calculation process, and the formula reference should be correct.

f. Ability to Analyze Business(AAB): Professional analysis of business needs and business types, such as products, markets, investments, finance, etc.

g. Clear Judgment(CJ): The answer should give a clear and directional judgment on the basic situation such as financial intention, financial sentiment, financial industry trend, and economic situation.

h. Safety Measure(SM): Financial and investment advice should have a certain degree of robustness and security, that is, to propose a safe and reliable investment strategy for users. Avoid risk-taking tendencies.

i. Financial Security and Compliance(FSC): The answer should conform to China's financial industry control system and China's market operation standards. For the financial industry and system interpretation standards that differ between China and other countries, the Chinese interpretation should be the only standard.

j. Risk Prediction and Control(RPC): The answer should make a clear judgment on the potential financial risks in the problem, and put forward the corresponding management and control strategies for the risks.

k. Financial Insight(FI): On the premise of making basic judgments on the situation and development of the financial industry, the opinions and suggestions put forward by the answers need to have a certain height and depth, which can see the essence through the problems, and then assist users to make the best judgment.
\begin{table}[H]
%\caption{Overall leaderboard. Closed-source models are in applicable leaderboard positions but not ranked.}

\centering
\begin{adjustbox}{scale=0.60}
\begin{tabular}{lccccc}
%{|X|c|c|c|c|}
\toprule
\centering
\textbf{Task type} & \textbf{Scoring standard} \\
\midrule
Fund Qualification & BR, IC, LI\\
Securities Qualification & BR, IC, LI\\
Banking Qualification & BR, IC, LI\\
Insurance Qualification (CICE) & BR, IC, LI\\
Economist & BR, IC, LI\\
Actuary & BR, IC, LI, MAE\\
Financial Planner & BR, LF, LI, SM\\
Futures Qualification & BR, IC, LI \\
Tax Accountant & BR, IC, LI\\
Certified Public Accountant (CPA) & BR, IC, LI \\
Financial Documents Summary & BR, LF, LI, FI \\
Financial Information Extraction & BR, IC \\
Financial Sentiment Judgment & BR, LI, CJ, FI \\
Mathematical Calculation & BR, IC, MAE \\
Logical Analysis & BR, LF, LI, CJ \\
Financial Compliance & BR, FSC, LI \\
Risk Management Application & BR, RPC, LI \\
Macroanalysis & BR, LF, LIy, CJ \\
Stock Analysis & BR, LF, LI, FI \\
Market Interpretation & BR, LF, LI, CJ \\
Financial Industry Analysis & BR, LF, LI, FI \\
Company Review & BR, LF, LI, AAB \\
Financial Report Comments & BR, LF, LI, AAB \\
Pan-investment Consultant Answer & BR, FSC, SM, FI \\
Fund Analysis & BR, FSC, SM, FI \\

\bottomrule
\end{tabular}
\end{adjustbox}
\caption{Correspondence between various tasks and corresponding evaluation criteria. BR: Basic Requirement, IC: Information Correctness, LF: Logic of Finance, LI: Language Intelligibility, MAE: Mathematical Ability and Efficiency, AAB: Ability to Analyze Business, CJ: Clear Judgment, SM: Safety Measure, FSC: Financial Security and Compliance, RPC: Risk Prediction and Control, FI: Financial Insight.}
\label{tab:model_Reliability}
\end{table}
\subsection{Assessment and Scoring Methodology}
Models are assessed through an API-based interaction, retrieving responses in either single or multi-turn conversational formats. For each question, the model's answer is scored according to the defined criteria. After obtaining scores for all tasks, these are normalized to a percentage scale, representing the overall performance of the model.

\subsection{Reliability Analysis}
To verify the reliability and practicality of the SC-Fin benchmark, a pre-evaluation experiment was conducted on four representative models, randomly selecting 120 questions per model from a pool of over 1000 questions. Reviewers evaluated the model outputs using a quality classification system with levels of excellence, satisfactory, passable, and failing. The results showed a high level of consistency and reliability in the assessment process.
\begin{table}[H]
%\caption{Overall leaderboard. Closed-source models are in applicable leaderboard positions but not ranked.}

\centering
\begin{adjustbox}{scale=0.80}
\begin{tabular}{lccccc}
%{|X|c|c|c|c|}
\toprule
\centering
\textbf{Model List} & \textbf{Reliability} \\
\midrule
Model One & 0.9084 \\
Model Two & 0.9924 \\
Model Three & 0.8760 \\
Model Four & 0.9695 \\
Average & 0.9366 \\

\bottomrule
\end{tabular}
\end{adjustbox}
\caption{Reliability Analysis}
\label{tab:model_Reliability}
\end{table}
\subsection{Grading Mechanism Explanation}

SuperCLUE-Fin adopts a tiered grading mechanism to classify models into three distinct tiers based on their overall performance. Tier 1 models must achieve a minimum grade of A, with at least one of their basic or applied capability grades being A or above, denoting their competence in fulfilling both consumer (ToC) and business (ToB) requirements. Tier 2 models must have grades no lower than B in both basic and applied capabilities, indicating they meet ToC needs but may require enhancements for ToB purposes. Tier 3 models, graded C or below, indicate significant room for improvement in both areas.

By implementing such a rigorous and multifaceted evaluation system, the SuperCLUE-Fin benchmark not only serves as a platform to measure the progress and identify the shortcomings of existing Chinese-native financial LLMs but also provides a roadmap for their future development, emphasizing the importance of improving database quality, refining the interpretation of financial problems based on domestic standards, and enhancing the models' analytical and decision-making capacities in line with China's financial context.
\begin{table}[H]
%\caption{Overall leaderboard. Closed-source models are in applicable leaderboard positions but not ranked.}

\centering
\begin{adjustbox}{scale=0.80}
\begin{tabular}{lccccc}
%{|X|c|c|c|c|}
\toprule
\centering
\textbf{Score Range} & \textbf{Level}  & \textbf{Tier} \\
\midrule
Above 75 & A+ & First Tier \\ 
70-75 & A & First Tier \\ 
65-70 & B & Second Tier \\ 
60-65 & C & Third Tier \\ 
Below 60 & D & Third Tier \\

\bottomrule
\end{tabular}
\end{adjustbox}
\caption{Grading Mechanism Explanation}
\label{tab:model_Reliability}
\end{table}
\subsection{Experimentation and Analysis}

The experimentation phase of the SuperCLUE-Fin benchmark involved a systematic evaluation of various financial large language models across a multitude of tasks designed to test their understanding, reasoning, and applicability in the financial domain. The analysis delves into the models' performances, strengths, weaknesses, and areas for improvement, providing valuable insights into the current state of domestic and international models in the context of Chinese financial services.
\begin{table}[H]
%\caption{Overall leaderboard. Closed-source models are in applicable leaderboard positions but not ranked.}
\centering
\begin{adjustbox}{scale=0.80}
\begin{tabular}{llcccc}
%{|X|c|c|c|c|}
\toprule
\centering
\textbf{Model Name} & \textbf{Organization} & \textbf{Access}  \\
\midrule
Baichuan2-13B-Chat & Baichuan & API \\
ChatGLM3-6B & ZhiPu & API \\
ERNIE Bot 4.0 & Baidu & API \\
Gemma-7b-instruct & Google & API \\
GLM-4 & ZhiPu & API \\
GPT-3.5 Turbo & OpenAI & API \\
GPT-4 & OpenAI & API \\
GPT-4 Turbo & OpenAI & API \\
MoonShot-v1-128K & MoonShot & API \\
qwen-finance-14B & Alibaba & API \\
SparkDesk V3.5 & Xunfei & API\\
\bottomrule
\end{tabular}
\end{adjustbox}
\caption{Model information}
\label{tab:model_information}
\end{table}

\subsection{Performance Analysis}
\begin{table}[H]
%\caption{Overall leaderboard. Closed-source models are in applicable leaderboard positions but not ranked.}
\centering
\begin{adjustbox}{scale=0.80}
\begin{tabular}{lccccc}
%{|X|c|c|c|c|}
\toprule
\centering
\textbf{Model Name} & \textbf{Model level}  \\
\midrule
GPT-4 Turbo & A+ \\
MoonShot-v1-128K & A \\
GLM-4 & A \\
SparkDesk V3.5 & B \\
ERNIE Bot 4.0 & B \\
GPT-4 & B \\
GPT-3.5 Turbo & C \\
Baichuan2-13B-Chat & C \\
ChatGLM3-6B & D \\
qwen-finance-14B & D \\
Gemma-7b-instruct & D \\
\bottomrule
\end{tabular}
\end{adjustbox}
\caption{SuperCLUE-Fin Review - Overall Ranking. Note: The same level models are sorted alphabetically.}
\label{tab:model_Reliability}
\end{table}
Upon assessment, it was observed that the overall financial capability of domestic models showed strong competitiveness, with GLM-4 and MoonShot-v1-128k reaching the highest tier with an A-grade, though still trailing behind the leading GPT-4 Turbo. Meanwhile, models like iFlytek StarFire V3.5 and Wenxin Yiyuan 4.0 achieved a B-grade, demonstrating better performance compared to GPT-4. However, the majority of the models fell within the C-tier, suggesting there is substantial room for advancement in the maturity and functionality of domestic financial LLMs.
\begin{table}[H]
%\caption{Overall leaderboard. Closed-source models are in applicable leaderboard positions but not ranked.}

\centering
\begin{adjustbox}{scale=0.65}
\begin{tabular}{lccccc}
%{|X|c|c|c|c|}
\toprule
\centering
\textbf{Model Name} & \textbf{Model level}  & \textbf{Basic Capability}  & \textbf{Applied Capability}\\
\midrule
GPT-4 Turbo & A+ & A+ & A+ \\
GLM-4 & A & A & B \\
MoonShot-v1-128K & A & A+ & B \\
GPT-4 & B & B & C \\
ERNIE Bot 4.0 & B & B & B \\
SparkDesk V3.5 & B & A & B \\
Baichuan2-13B-Chat & C & C & C \\
GPT-3.5 Turbo & C & B & C \\
ChatGLM3-6B & D & D & C \\
Gemma-7b-instruct & D & D & D \\
Average Level & B & B & B \\

\bottomrule
\end{tabular}
\end{adjustbox}
\caption{SuperCLUE-Fin Review - Summary of each ability level. Note: The same level models are sorted alphabetically.}
\label{tab:model_Reliability}
\end{table}
The analysis further revealed disparities between the models' basic and applied capabilities. While they generally excelled in basic financial knowledge (such as financial theory, market regulations, and common financial instruments), there was a notable deficiency in their ability to apply this knowledge in more complex, real-world scenarios, requiring strategic decision-making, risk assessment, and compliance considerations.
\begin{table}[H]
%\caption{Overall leaderboard. Closed-source models are in applicable leaderboard positions but not ranked.}

\centering
\begin{adjustbox}{scale=0.80}
\begin{tabular}{lcccccc}
%{|X|c|c|c|c|}
\toprule
\centering
\textbf{Model Name} & \textbf{FF}  & \textbf{FIEA}  & \textbf{MCLA} & \textbf{FCRM} & \textbf{IR} &\textbf{IA}\\
\midrule
GPT-4 Turbo & A+ & A+ & A+ & A+ & A & A\\
GLM-4 & A+ & B & A & A & B & A \\
MoonShot-v1-128K & A+ & B & B & A & B & B\\
SparkDesk V3.5 & A & B & A & A & B & B\\
ERNIE Bot 4.0 & A & B & C & A & B & B\\
GPT-4 & A & B & B & B & C & B\\
GPT-3.5 Turbo & B & D & C & B & C & C\\
Baichuan2-13B-Chat & B & D & D & B & C & C\\
ChatGLM3-6B & C & D & D & B & C & C\\
qwen-finance-14B & C & C & D & C & D & C\\
Gemma-7b-instruct & D & C & D & C & D & D\\

\bottomrule
\end{tabular}
\end{adjustbox}
\caption{SuperCLUE-Fin Review - Summary of each task level (primary class). Note: The same level models are sorted alphabetically.FF: Financial Fundamentals, FIEA: Financial Information Extraction and Analysis, MCLA: Mathematical Calculation and Logical Analysis, FCRM: Financial Compliance and Risk Management, IR: Investment Research, IA: Investment Adviser.}
\label{tab:model_Reliability}
\end{table}
In the specific task categories, models demonstrated varying degrees of success. For instance, in the financial understanding and cognition tasks, models could effectively summarize financial documents and extract pertinent information. However, in the realm of financial number-theoretic calculations and logical analysis, discrepancies became evident. The models' ability to handle complex financial computations and to provide detailed, contextually accurate explanations varied significantly, sometimes failing to account for potential risks or limitations inherent in financial decisions, such as stock buybacks.

Taking the correlation Analysis between Financial Fundamentals and the tasks of Mathematical Calculation and Logical Analysis as an example, as shown in the figure below, the model shows a high correlation between these two tasks. Without considering the impact of the model's computing ability on the final result, the higher the model's mastery of financial knowledge is, the stronger the model's ability to solve the same financial mathematical calculation problems is. This shows that it is necessary to improve the quality of large financial model database and improve the interpretation standard of models for financial problems to improve the performance of large financial models.
\begin{figure}[h]
\centering
%\raggedright
% [width=0.7\textwidth, height=0.6\textheight]
\includegraphics[width=0.5\textwidth, height=0.2\textheight]{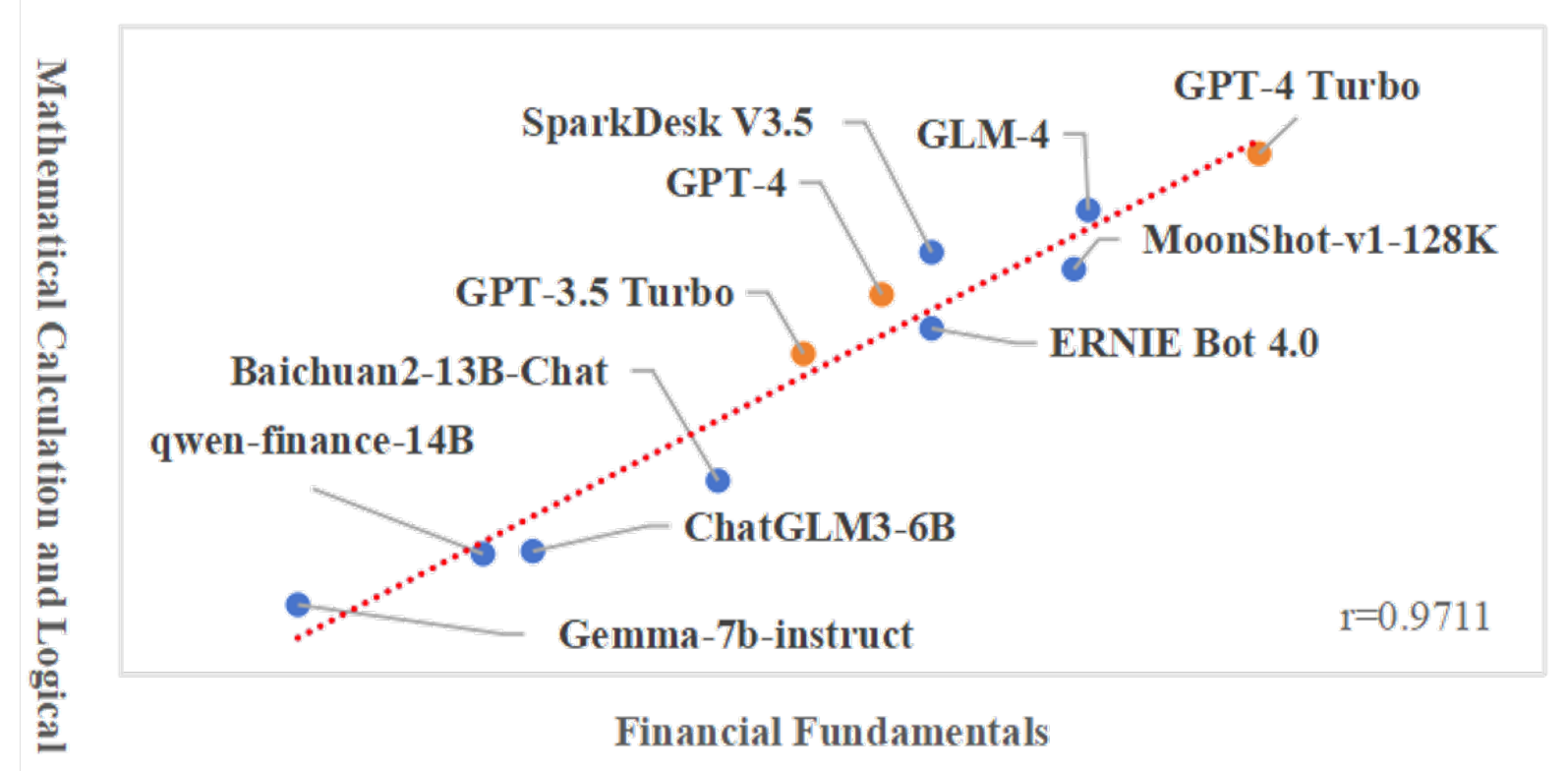}
\caption{Financial Fundamentals - Mathematical Calculation and Logical Analysis}
\label{fig:better_performance_with_response_length_increasing}
\end{figure}

\subsection{Performance analysis of each task}

1)Financial Fundamentals

In the part of Financial Fundamentals, the grade of most models can reach B or above, which indicates that the models have a good grasp of basic knowledge in the financial field, and have the foundation to further expand the large model to deal with more complex financial problems and financial business.
\begin{table}[H]
%\caption{Overall leaderboard. Closed-source models are in applicable leaderboard positions but not ranked.}

\centering
\begin{adjustbox}{scale=0.55}
\begin{tabular}{lccccccccccc}
%{|X|c|c|c|c|}
\toprule
\centering
\textbf{Model Name} & \textbf{Model Level}  & \textbf{FdQ}  & \textbf{SQ} & \textbf{BQ} & \textbf{CICE} &\textbf{Ec} &\textbf{Ac} &\textbf{FP} &\textbf{FsQ} &\textbf{TA} &\textbf{CPA}\\
\midrule
GLM-4 & A+ & A+ & A+ & A+ & A+ & A+ &B & A & A & A+ & A+ \\
GPT-4 Turbo & A+ & A+ & A+ & A+ & A+ & A+ & A+ & A+ & A+ & A+ &A+\\
MoonShot-v1-128K & A+ & A+ & A+ & A+ & A+ & A+ &B & A & A+ & A+ & A+ \\
GPT-4 & A & A & A & A & A & A & B & B & A+ & A & B \\
ERNIE Bot 4.0 & A & A & A+ & A+ & A+ & A & C & A & A+ & A & B \\
SparkDesk V3.5 & A & A & A+ & A+ & A & A & B & A & A+ & B & B \\
Baichuan2-13B-Chat & B & B & B & A+ & B & A & D & B & C & A & B \\
GPT-3.5 Turbo & B & B & A & A+ & A & B & D & B & A+ & B & B \\
ChatGLM3-6B & C & C & C & A & B & D & D & C & C & C & C \\
qwen-finance-14B & C & C & C & C & B & D & D & C & D & D & C \\
Gemma-7b-instruct & D & D & D & C & D & D & D & D & D & D & D \\

\bottomrule
\end{tabular}
\end{adjustbox}
\caption{Financial Fundamentals Overview of model results. Note: The same level models are sorted alphabetically.FdQ: Fund Qualification, SQ: Securities Qualification, BQ: Banking Qualification, CICE: Insurance Qualification (CICE), Ec: Economist, Ac: Actuary, FP: Financial Planner, FsQ: Futures Qualification, TA: Tax Accountant, CPA: Certified Public Accountant (CPA)}
\label{tab:model_Reliability}
\end{table}
2)Financial Information Extraction and Analysis
In the part of Financial Information Extraction and Analysis, except GPT-4 Turbo, the overall grade of other models is B or below, which indicates that there are still big problems in dealing with financial information in the financial large model. For example, it is not sensitive enough to capture financial related words, not enough to analyze financial professional words, and not enough to summarize and summarize financial terms skillfully.
\begin{table}[H]
%\caption{Overall leaderboard. Closed-source models are in applicable leaderboard positions but not ranked.}

\centering
\begin{adjustbox}{scale=0.8}
\begin{tabular}{lcccc}
%{|X|c|c|c|c|}
\toprule
\centering
\textbf{Model Name} & \textbf{Model Level}  & \textbf{FDS}  & \textbf{FIE} & \textbf{FSJ} \\
\midrule
GPT-4 Turbo & A+ & A & A+ & A+ \\
GLM-4 & B & B & B & B \\
GPT-4 & B & C & A+ & B \\
MoonShot-v1-128K & B & B & B & B \\
ERNIE Bot 4.0 & B & B & A & B \\
SparkDesk V3.5 & B & B & B & B \\
Gemma-7b-instruct & C & D & C & C \\
qwen-finance-14B & C & C & D & D \\
Baichuan2-13B-Chat & D & C & D & C \\
ChatGLM3-6B & D & C & D & B \\
GPT-3.5 Turbo & D & C & D & C \\

\bottomrule
\end{tabular}
\end{adjustbox}
\caption{Financial Information Extraction and Analysis Overview of model results. Note: The same level models are sorted alphabetically.FDS: Financial Documents Summary, FIE: Financial Information Extraction, FSJ: Financial Sentiment Judgment}
\label{tab:model_Reliability}
\end{table}
3)Mathematical Calculation and Logical Analysis
In the part of Mathematical Calculation and Logical Analysis, all models show strong polarization phenomenon, that is, the model with higher comprehensive grade also shows better performance in mathematical calculation and logical analysis ability. However, the model with low comprehensive level showed significantly lower performance than the comprehensive ability in the above two types of tasks.
\begin{table}[H]
%\caption{Overall leaderboard. Closed-source models are in applicable leaderboard positions but not ranked.}

\centering
\begin{adjustbox}{scale=0.8}
\begin{tabular}{lccc}
%{|X|c|c|c|c|}
\toprule
\centering
\textbf{Model Name} & \textbf{Model Level}  & \textbf{MC}  & \textbf{LA} \\
\midrule
GPT-4 Turbo & A+ & A+ & A+ \\
GLM-4 & A & A & A \\
SparkDesk V3.5 & A & B & A \\
GPT-4 & B & B & B \\
MoonShot-v1-128K & B & B & A \\
GPT-3.5 Turbo & C & C & C \\
ERNIE Bot 4.0 & C & D & A \\
Baichuan2-13B-Chat & D & D & B \\
ChatGLM3-6B & D & D & C \\
Gemma-7b-instruct & D & D & D \\
qwen-finance-14B & D & D & D \\

\bottomrule
\end{tabular}
\end{adjustbox}
\caption{Mathematical Calculation and Logical Analysis Overview of model results. Note: The same level models are sorted alphabetically.MC: Mathematical Calculation, LA:  Logical Analysis}
\label{tab:model_Reliability}
\end{table}
4)Financial Compliance and Risk Management
In the part of Financial Compliance and Risk Management, each model almost shows no lower than the comprehensive level, which indicates that the financial model has a relatively mature ability in dealing with the problems related to laws and regulations and risk aversion and management. This is the key factor to ensure that the models do not break the law as they continue to evolve.
\begin{table}[H]
%\caption{Overall leaderboard. Closed-source models are in applicable leaderboard positions but not ranked.}

\centering
\begin{adjustbox}{scale=0.8}
\begin{tabular}{lccc}
%{|X|c|c|c|c|}
\toprule
\centering
\textbf{Model Name} & \textbf{Model Level}  & \textbf{FC}  & \textbf{RMA} \\
\midrule
GPT-4 Turbo & A+ & A & A+ \\
GLM-4 & A & A & A \\
MoonShot-v1-128K & A & A & A \\
ERNIE Bot 4.0 & A & A & A \\
SparkDesk V3.5 & A & A & A \\
Baichuan2-13B-Chat & B & B & B \\
ChatGLM3-6B & B & B & B \\
GPT-3.5 Turbo & B & B & B \\
GPT-4 & B & B & B \\
Gemma-7b-instruct & C & D & C \\
qwen-finance-14B & C & C & C \\

\bottomrule
\end{tabular}
\end{adjustbox}
\caption{Financial Compliance and Risk Management Overview of model results. Note: The same level models are sorted alphabetically.FC: Financial Compliance, RMA: Risk Management Application}
\label{tab:model_Reliability}
\end{table}
5)Investment Research
In the Investment Research section, the models almost all performed below their composite grade. Considering the specific content involved in this task type, the above phenomenon indicates that the ability of the financial model to deal with the market, the industry and the specific business within the company is not good, and it is expected that the task ability will be further improved with the further development of the model.
\begin{table}[H]
%\caption{Overall leaderboard. Closed-source models are in applicable leaderboard positions but not ranked.}

\centering
\begin{adjustbox}{scale=0.6}
\begin{tabular}{lccccccc}
%{|X|c|c|c|c|}
\toprule
\centering
\textbf{Model Name} & \textbf{Model Level}  & \textbf{Ma}  & \textbf{SA}  & \textbf{MI} & \textbf{FIA} & \textbf{CR} & \textbf{FRC}\\
\midrule
GPT-4 Turbo & A & A & D & A & B & A & A \\
GLM-4 & B & A & D & A & B & B & A \\
MoonShot-v1-128K & B & A & D & A & B & B & A \\
ERNIE Bot 4.0 & B & A & D & B & D & B & A \\
SparkDesk V3.5 & B & B & D & A & C & B & C \\
Baichuan2-13B-Chat & C & B & D & B & C & C & C \\
ChatGLM3-6B & C & B & D & C & D & D & D \\
GPT-3.5 Turbo & C & C & D & C & D & C & D \\
GPT-4 & C & B & D & B & C & C & C \\
Gemma-7b-instruct & D & D & D & D & D & D & D \\
qwen-finance-14B & D & C & D & D & D & D & D \\

\bottomrule
\end{tabular}
\end{adjustbox}
\caption{Investment Research Overview of model results. Note: The same level models are sorted alphabetically.Ma: Macroanalysis, SA:  Stock Analysis, MI: Market Interpretation, FIA: Financial Industry Analysis, CR: Company Review, FRC: Financial Report Comments}
\label{tab:model_Reliability}
\end{table}
6)Investment Adviser
In the Investment Adviser section, the performance of each model is similar to that in the Investment Research section, that is, the performance is not satisfactory when specific business QA is involved. In particular, when simulated investment advisers answer users' investment-related questions, they are not in-depth and detailed enough, and their suggestions are relatively vague.
\begin{table}[H]
%\caption{Overall leaderboard. Closed-source models are in applicable leaderboard positions but not ranked.}

\centering
\begin{adjustbox}{scale=0.8}
\begin{tabular}{lccc}
%{|X|c|c|c|c|}
\toprule
\centering
\textbf{Model Name} & \textbf{Model Level}  & \textbf{PCA}  & \textbf{FA}\\
\midrule
GLM-4 & A & A & A+ \\
GPT-4 Turbo & A & A & B \\
GPT-4 & B & B & B \\
MoonShot-v1-128K & B & B & A \\
ERNIE Bot 4.0 & B & B & A \\
SparkDesk V3.5 & B & B & B \\
Baichuan2-13B-Chat & C & C & D \\
ChatGLM3-6B & C & C & C \\
GPT-3.5 Turbo & C & C & C \\
qwen-finance-14B & C & C & D \\
Gemma-7b-instruct & D & D & D \\

\bottomrule
\end{tabular}
\end{adjustbox}
\caption{Investment Adviser Overview of model results. Note: The same level models are sorted alphabetically.PCA: Pan-investment Consultant Answer, FA: Fund Analysis}
\label{tab:model_Reliability}
\end{table}
\subsection{Reliability and Validity}
The reliability of the SuperCLUE-Fin benchmark was rigorously examined by conducting experiments on four representative models, with human evaluators reviewing a random subset of responses. The high rate of agreement between the model outputs and evaluator judgments confirmed the reliability of the assessment method. The scoring system was shown to be consistent and fair, with a clear distinction between model performance levels.
From the empirical data and analysis, it can be concluded that although certain domestic models have reached commendable levels of competency, there is still a gap to bridge before attaining the same level of sophistication and adaptability as seen in top-performing international models like GPT-4 Turbo. Specifically, the specialized capabilities of financial LLMs in China need considerable enhancement, particularly in relation to application-oriented tasks and the provision of contextually precise, balanced, and risk-aware recommendations.
It was also noted that the relationship between a model's mastery of financial knowledge and its performance in financial calculations underscored the importance of having a robust financial knowledge base and adherence to consistent and accurate interpretations of financial principles, especially those aligned with Chinese financial regulations.
Overall, the SuperCLUE-Fin benchmark not only serves as a powerful diagnostic tool to evaluate the strengths and weaknesses of financial LLMs but also provides a roadmap for future model development, emphasizing the necessity to upgrade databases, refine interpretations of financial problems, and bolster the models' ability to navigate the intricacies of the Chinese financial landscape. This will ultimately lead to more effective, compliant, and customer-centric financial services powered by AI technologies.

\section{Related Work}
In the field of large language models (LLMs) and their application in the financial domain, several benchmarks and models have emerged globally. The FINQA dataset, is a large-scale collection of financial reports paired with expert-annotated questions and answers, aiming to enhance the automation of financial data analysis through complex numerical reasoning (Chen et al., 2021).

Another relevant contribution comes from the Financial Narrative Processing (FNP) shared task series, which has hosted annual competitions since 2018. The tasks involve extracting and summarizing financial information from earnings calls and annual reports, sentiment analysis, and event extraction (Chen et al., 2021). These efforts have significantly pushed the boundaries of natural language processing (NLP) techniques in the financial sector and provided valuable insights into the performance and limitations of contemporary NLP models in handling complex financial text.

On the model side, OpenAI's GPT series, especially GPT-3.5, GPT-4 have demonstrated impressive results when fine-tuned for financial applications. They have been used for generating financial reports, giving investment advice, and assisting in compliance tasks (OpenAI et al., 2023). However, these models were primarily developed for a global English-speaking audience and might not fully capture the specificities and complexities of the Chinese financial market.

Our proposed SuperCLUE-Fin benchmark builds upon these prior works by tailoring the evaluation to the Chinese financial context. Unlike previous benchmarks, SC-Fin emphasizes native Chinese content and financial regulations, aiming to address the unique challenges faced by AI models operating within China's financial ecosystem. It incorporates a diverse range of tasks, from fundamental financial knowledge to complex applications, ensuring that the tested models can not only comprehend and analyze financial data but also adhere to local regulatory guidelines and cultural nuances, thus paving the way for more accurate, reliable, and contextually relevant financial AI solutions in China.

% BELOW IS OLD FROM MATH6:
% Benchmarks to evaluate the reasoning skills of language models have gained increasing research attention. Existing datasets mostly focus on English, including GSM8K for mathematical reasoning~\citep{[2]}, MATH for complexity mathematical problem ~\citep{[3]}.  For general LLMs benchmark, we can find MT-bench ~\citep{[4]}, AlpacaEval ~\citep{[5]}. We can find reasoning benchmarks for NLP, such as  WinoGrad Schema Challenge for commonsense reasoning ~\citep{[6]}, and ARC for scientific question answering ~\citep{[7]}.
% Our work aims to close the gap for the Chinese through a systematically designed mathematical reasoning benchmark. Our focus is to provide a benchmark to evaluate the general reasoning skills of Chinese language models. The diverse problems and reasoning patterns in SC-Math6 complement these methods to inspire new model designs and training strategies targeting enhanced mathematical intelligence.

% category figure here

% ABC...
\section{Conclusion}
The SC-Fin benchmark represents a significant step towards a more standardized, professional, and transparent assessment process for FLMs in the Chinese financial domain. Its rigorous testing methodology and scoring system not only provide a comprehensive measure of the current state of these models but also offer valuable insights for developers to improve upon. The introduction of a tiered grading mechanism further enhances the credibility of the assessment and aids stakeholders in gauging the suitability of different models for various use cases.

\appendix

\section{Examples for problem in SC-Fin}
Below are examples of problems in SuperCLUE-Fin.

\subsection{Basic Capability}

\subsection{Example 1: Fund Qualification}

Question: The net value of QDII fund issues regulation: QDII fund has any special requirements in net value issues?

Follow-up: What is the impact of these regulations on investor transparency and confidence?

\subsection{Example 2: Actuary}

Question: An insurance company uses claims data to predict future health insurance costs. If the average claim rate over the past five years is 0.025, the average claim amount is 1,500 dollars, and the claim rate is expected to rise by 0.1 in the future, how should the company adjust its premium?

Follow-up: If claims are expected to rise by 0.2, how much do premiums need to adjust to account for higher costs?

\subsection{Example 3: Financial Planner}

Question: How to optimize the portfolio to enhance its resilience to risk?

Follow-up: Please explain in detail the difference between asset allocation and diversification and their importance.

\subsection{Example 4: Financial Documents Summary}

Question: Please answer the question based on the information you already know. The following is a summary summary. Requirements: no more than 150 words, focus, organized content.

Known information: Amperon (301413) is mainly engaged in temperature and pressure sensors, and the downstream is mainly automotive electronics and home appliances. The company's product line includes thermistors and temperature sensors, oxygen sensors, pressure sensors, downstream applications are mainly in automotive, home appliances, industrial energy storage, etc. In recent years, the proportion of pressure sensor revenue has increased significantly.

... < middle omission >...

In addition, the company has obtained a number of MEMS pressure sensor projects from a well-known European oems customer, some of which have realized SOP in the first quarter of 2024.

\subsection{Example 5: Mathematical Calculation}

Question: Portfolio diversification: Xiaohong invests in four different stocks, and the investment proportion and expected return rate of each stock are 0.3 and 0.05 for A stock, 0.3 and 0.06 for B stock, 0.2 and 0.07 for C stock, and 0.2 and 0.08 for D stock respectively. The expected return rate of her portfolio is calculated.

Follow-up: If the expected return on A-stock falls to 0.03, how will the expected return on the entire portfolio adjust?

\subsection{Example 6: Logical Analysis}

Question: The ROE of a listed company is higher than 0.25 for three consecutive years, but its P/E ratio is lower than the industry average. The company's management believed that the market was undervaluing the company and decided to initiate a share buyback program. Q. What is the logic behind the company's move? Can share repurchases effectively boost stock prices?

Follow-up: What other factors should be considered in this decision?

\subsection{Applied Capabilities}

\subsection{Example 7: Financial Compliance}

Question: What are the regulatory policies for high-frequency trading in the securities market?

Follow-up: How do these policies affect market fairness and transparency?

\subsection{Example 8: Financial Industry Analysis}

Question: Answer the question based on known information.
Known information: Title: Coal mining industry tracking weekly report: off-season demand continues to be weak coal prices stabilize and shock. Source: Soochow Securities.

Content: Industry status: thermal coal, off-season demand remains weak, port coal prices stabilize and shock. This week (April 7 to April 12), the spot price of thermal coal at the port fell by 15 yuan/ton month-on-month to 801 yuan/ton.
... < middle omission >...
We maintain the "overweight" rating of the industry, recommend high flexibility targets China Grand Energy and China Grand Logistics, and suggest paying attention to Haohua Energy; In addition, we still recommend insurance OCI capital allocation of high dividend targets: China Shenhua, Shaanxi Coal Industry, it is suggested to pay attention to Yankuang Energy. Risk warning: downstream demand is less than expected; Coal prices fell sharply as supply protection was stronger than expected.
Excuse me, based on the current market trend, why the continued optimism for coal prices in the first half of 2024?
Follow-up: What is the current supply and demand situation of the coal market and how does it affect the trend of coal prices?

\subsection{Example 9: Company Review}

Question: Answer the question based on known information.
The following is the content of the article:
On April 11, Yum China (HK 09987, HK 299.8, market value HK 117.9 billion) released an open letter to shareholders, in which CEO Qu Cuirong mentioned that the company plans to accelerate the pace of returning at least US 3 billion to shareholders in the next three years
... < middle omission >...
At the same time, our flexible store model helps us reduce our upfront investment costs, and our strong own supply chain management capabilities and logistics network help us reach remote areas, all of which give us a strong competitive advantage.
Please, Yum China's operating profit in 2023 will reach 1.1 billion US dollars, with core operating profit growing 0.79. Based on the company's store expansion plan and lower-tier city layout in the next three years, how do you think its profitability and valuation are reasonable? What are the advantages over other companies in the same industry?
Follow-up: What do the frequent debt problems of Beingmate Group reflect the deficiencies in corporate governance and fund management? What improvement measures does the company need in financing and investment decisions in the future?

\subsection{Example 10: Pan-investment Consultant Answer}

Question: Futures investment: recently the price of bulk commodities fluctuates greatly, how should individual investors participate in the futures market?

Follow-up: Compared with securities investment, what are the characteristics of futures investment in terms of leverage ratio and trading mechanism? What are the risks and benefits?
\label{appendix:appendix_a}


\begin{thebibliography}{34}
\providecommand{\natexlab}[1]{#1}

\bibitem[{Brown et~al.(2020)Brown, Mann, Ryder, Subbiah, Kaplan, Dhariwal,
  Neelakantan, Shyam, Sastry, Askell, Agarwal, Herbert-Voss, Krueger, Henighan,
  Child, Ramesh, Ziegler, Wu, Winter, Hesse, Chen, Sigler, Litwin, Gray, Chess,
  Clark, Berner, McCandlish, Radford, Sutskever, and Amodei}]{brown2020gpt}
Brown, T.~B.; Mann, B.; Ryder, N.; Subbiah, M.; Kaplan, J.; Dhariwal, P.;
  Neelakantan, A.; Shyam, P.; Sastry, G.; Askell, A.; Agarwal, S.;
  Herbert-Voss, A.; Krueger, G.; Henighan, T.; Child, R.; Ramesh, A.; Ziegler,
  D.~M.; Wu, J.; Winter, C.; Hesse, C.; Chen, M.; Sigler, E.; Litwin, M.; Gray,
  S.; Chess, B.; Clark, J.; Berner, C.; McCandlish, S.; Radford, A.; Sutskever,
  I.; and Amodei, D. 2020.
\newblock Language Models are Few-Shot Learners.
\newblock arXiv:2005.14165.

\bibitem[{Chiang et~al.(2023)Chiang, Li, Lin, Sheng, Wu, Zhang, Zheng, Zhuang,
  Zhuang, Gonzalez, Stoica, and Xing}]{vicuna2023}
Chiang, W.-L.; Li, Z.; Lin, Z.; Sheng, Y.; Wu, Z.; Zhang, H.; Zheng, L.;
  Zhuang, S.; Zhuang, Y.; Gonzalez, J.~E.; Stoica, I.; and Xing, E.~P. 2023.
\newblock Vicuna: An Open-Source Chatbot Impressing GPT-4 with 90\%* ChatGPT
  Quality.

\bibitem[{Devlin et~al.(2019)Devlin, Chang, Lee, and
  Toutanova}]{devlin-etal-2019-bert}
Devlin, J.; Chang, M.-W.; Lee, K.; and Toutanova, K. 2019.
\newblock {BERT}: Pre-training of Deep Bidirectional Transformers for Language
  Understanding.
\newblock In \emph{Proceedings of the 2019 Conference of the North {A}merican
  Chapter of the Association for Computational Linguistics: Human Language
  Technologies, Volume 1 (Long and Short Papers)}, 4171--4186. Minneapolis,
  Minnesota: Association for Computational Linguistics.

\bibitem[{Dolan and Brockett(2005)}]{dolan-brockett-2005-MRPC}
Dolan, W.~B.; and Brockett, C. 2005.
\newblock Automatically Constructing a Corpus of Sentential Paraphrases.
\newblock In \emph{Proceedings of the Third International Workshop on
  Paraphrasing ({IWP}2005)}.

\bibitem[{Du et~al.(2022)Du, Qian, Liu, Ding, Qiu, Yang, and Tang}]{du2022glm}
Du, Z.; Qian, Y.; Liu, X.; Ding, M.; Qiu, J.; Yang, Z.; and Tang, J. 2022.
\newblock GLM: General Language Model Pretraining with Autoregressive Blank
  Infilling.
\newblock In \emph{Proceedings of the 60th Annual Meeting of the Association
  for Computational Linguistics (Volume 1: Long Papers)}, 320--335.

\bibitem[{Goyal, Li, and Durrett(2023)}]{goyal2023gpt-sum}
Goyal, T.; Li, J.~J.; and Durrett, G. 2023.
\newblock News Summarization and Evaluation in the Era of GPT-3.
\newblock arXiv:2209.12356.

\bibitem[{Gu et~al.(2023)Gu, Zhu, Ye, Zhang, Wang, Jiang, Xiong, Li, He, Xu,
  Huang, Wang, Wang, Zheng, Feng, and Xiao}]{gu2023xiezhi}
Gu, Z.; Zhu, X.; Ye, H.; Zhang, L.; Wang, J.; Jiang, S.; Xiong, Z.; Li, Z.; He,
  Q.; Xu, R.; Huang, W.; Wang, Z.; Wang, S.; Zheng, W.; Feng, H.; and Xiao, Y.
  2023.
\newblock Xiezhi: An Ever-Updating Benchmark for Holistic Domain Knowledge
  Evaluation.
\newblock arXiv:2306.05783.

\bibitem[{Hendrycks et~al.(2020)Hendrycks, Burns, Basart, Zou, Mazeika, Song,
  and Steinhardt}]{Dan2020mmlu}
Hendrycks, D.; Burns, C.; Basart, S.; Zou, A.; Mazeika, M.; Song, D.; and
  Steinhardt, J. 2020.
\newblock Measuring Massive Multitask Language Understanding.
\newblock \emph{CoRR}, abs/2009.03300.

\bibitem[{Huang et~al.(2023)Huang, Bai, Zhu, Zhang, Zhang, Su, Liu, Lv, Zhang,
  Lei, Fu, Sun, and He}]{huang2023ceval}
Huang, Y.; Bai, Y.; Zhu, Z.; Zhang, J.; Zhang, J.; Su, T.; Liu, J.; Lv, C.;
  Zhang, Y.; Lei, J.; Fu, Y.; Sun, M.; and He, J. 2023.
\newblock C-Eval: A Multi-Level Multi-Discipline Chinese Evaluation Suite for
  Foundation Models.
\newblock \emph{arXiv preprint arXiv:2305.08322}.

\bibitem[{Kasneci et~al.(2023)Kasneci, Seßler, Küchemann, Bannert,
  Dementieva, Fischer, Gasser, Groh, Günnemann, Hüllermeier, and
  et~al.}]{kasneci2023education_survey}
Kasneci, E.; Seßler, K.; Küchemann, S.; Bannert, M.; Dementieva, D.; Fischer,
  F.; Gasser, U.; Groh, G.; Günnemann, S.; Hüllermeier, E.; and et~al. 2023.
\newblock ChatGPT for Good? On Opportunities and Challenges of Large Language
  Models for Education.

\bibitem[{Lee~Rodgers and Nicewander(1988)}]{lee1988pearson}
Lee~Rodgers, J.; and Nicewander, W.~A. 1988.
\newblock Thirteen ways to look at the correlation coefficient.
\newblock \emph{The American Statistician}, 42(1): 59--66.

\bibitem[{Lewkowycz et~al.(2022)Lewkowycz, Slone, Andreassen, Freeman, Dyer,
  Mishra, Gur-Ari, Lee, Sohl-dickstein, Chiafullo, Fedus, Fiedel, Liu, Misra,
  and Ramasesh}]{Srivastava2022BIGbench}
Lewkowycz, A.; Slone, A.; Andreassen, A.; Freeman, D.; Dyer, E.~S.; Mishra, G.;
  Gur-Ari, G.; Lee, J.; Sohl-dickstein, J.; Chiafullo, K.; Fedus, L.~B.;
  Fiedel, N.; Liu, R.; Misra, V.; and Ramasesh, V.~V. 2022.
\newblock Beyond the Imitation Game: Quantifying and extrapolating the
  capabilities of language models.
\newblock Technical report.

\bibitem[{Li et~al.(2023)Li, Zhang, Dubois, Taori, Gulrajani, Guestrin, Liang,
  and Hashimoto}]{alpaca_eval}
Li, X.; Zhang, T.; Dubois, Y.; Taori, R.; Gulrajani, I.; Guestrin, C.; Liang,
  P.; and Hashimoto, T.~B. 2023.
\newblock AlpacaEval: An Automatic Evaluator of Instruction-following Models.
\newblock \url{https://github.com/tatsu-lab/alpaca_eval}.

\bibitem[{Liang et~al.(2022)Liang, Bommasani, Lee, Tsipras, Soylu, Yasunaga,
  Zhang, Narayanan, Wu, Kumar, Newman, Yuan, Yan, Zhang, Cosgrove, Manning,
  Ré, Acosta-Navas, Hudson, Zelikman, Durmus, Ladhak, Rong, Ren, Yao, Wang,
  Santhanam, Orr, Zheng, Yuksekgonul, Suzgun, Kim, Guha, Chatterji, Khattab,
  Henderson, Huang, Chi, Xie, Santurkar, Ganguli, Hashimoto, Icard, Zhang,
  Chaudhary, Wang, Li, Mai, Zhang, and Koreeda}]{liang2022helm}
Liang, P.; Bommasani, R.; Lee, T.; Tsipras, D.; Soylu, D.; Yasunaga, M.; Zhang,
  Y.; Narayanan, D.; Wu, Y.; Kumar, A.; Newman, B.; Yuan, B.; Yan, B.; Zhang,
  C.; Cosgrove, C.; Manning, C.~D.; Ré, C.; Acosta-Navas, D.; Hudson, D.~A.;
  Zelikman, E.; Durmus, E.; Ladhak, F.; Rong, F.; Ren, H.; Yao, H.; Wang, J.;
  Santhanam, K.; Orr, L.; Zheng, L.; Yuksekgonul, M.; Suzgun, M.; Kim, N.;
  Guha, N.; Chatterji, N.; Khattab, O.; Henderson, P.; Huang, Q.; Chi, R.; Xie,
  S.~M.; Santurkar, S.; Ganguli, S.; Hashimoto, T.; Icard, T.; Zhang, T.;
  Chaudhary, V.; Wang, W.; Li, X.; Mai, Y.; Zhang, Y.; and Koreeda, Y. 2022.
\newblock Holistic Evaluation of Language Models.
\newblock arXiv:2211.09110.

\bibitem[{Nov, Singh, and Mann(2023)}]{nov2023chatgpt_medical}
Nov, O.; Singh, N.; and Mann, D. 2023.
\newblock Putting ChatGPT's Medical Advice to the (Turing) Test.
\newblock arXiv:2301.10035.

\bibitem[{OpenAI(2023)}]{openai2023gpt4}
OpenAI. 2023.
\newblock GPT-4 Technical Report.
\newblock arXiv:2303.08774.

\bibitem[{Ouyang et~al.(2022)Ouyang, Wu, Jiang, Almeida, Wainwright, Mishkin,
  Zhang, Agarwal, Slama, Ray, Schulman, Hilton, Kelton, Miller, Simens, Askell,
  Welinder, Christiano, Leike, and Lowe}]{ouyang2022RLHF}
Ouyang, L.; Wu, J.; Jiang, X.; Almeida, D.; Wainwright, C.~L.; Mishkin, P.;
  Zhang, C.; Agarwal, S.; Slama, K.; Ray, A.; Schulman, J.; Hilton, J.; Kelton,
  F.; Miller, L.; Simens, M.; Askell, A.; Welinder, P.; Christiano, P.; Leike,
  J.; and Lowe, R. 2022.
\newblock Training language models to follow instructions with human feedback.
\newblock arXiv:2203.02155.

\bibitem[{Rajpurkar et~al.(2016)Rajpurkar, Zhang, Lopyrev, and
  Liang}]{rajpurkar2016squad}
Rajpurkar, P.; Zhang, J.; Lopyrev, K.; and Liang, P. 2016.
\newblock Squad: 100,000+ questions for machine comprehension of text.
\newblock \emph{arXiv preprint arXiv:1606.05250}.

\bibitem[{Sallam(2023)}]{sallam2023chatgpt_in_healthcare}
Sallam, M. 2023.
\newblock ChatGPT utility in healthcare education, research, and practice:
  systematic review on the promising perspectives and valid concerns.
\newblock In \emph{Healthcare}, volume~11, 887. MDPI.

\bibitem[{Sarlin et~al.(2019)Sarlin, DeTone, Malisiewicz, and
  Rabinovich}]{paul2019superglue}
Sarlin, P.; DeTone, D.; Malisiewicz, T.; and Rabinovich, A. 2019.
\newblock SuperGlue: Learning Feature Matching with Graph Neural Networks.
\newblock \emph{CoRR}, abs/1911.11763.

\bibitem[{Socher et~al.(2013)Socher, Perelygin, Wu, Chuang, Manning, Ng, and
  Potts}]{socher2013SST}
Socher, R.; Perelygin, A.; Wu, J.; Chuang, J.; Manning, C.~D.; Ng, A.~Y.; and
  Potts, C. 2013.
\newblock Recursive deep models for semantic compositionality over a sentiment
  treebank.
\newblock In \emph{Proceedings of the 2013 conference on empirical methods in
  natural language processing}, 1631--1642.

\bibitem[{Spearman(1987)}]{spearman1987spearman}
Spearman, C. 1987.
\newblock The Proof and Measurement of Association between Two Things.
\newblock \emph{The American Journal of Psychology}, 100(3/4): 441--471.

\bibitem[{Sun et~al.(2023)Sun, Zhang, He, Li, Cheng, Yan, Liu, Shao, Tang,
  Zhao, Chen, Zheng, Zhou, Li, Zhan, Zhou, Li, Yang, Wu, Yin, Huang, and
  Qiu}]{sun2023moss}
Sun, T.; Zhang, X.; He, Z.; Li, P.; Cheng, Q.; Yan, H.; Liu, X.; Shao, Y.;
  Tang, Q.; Zhao, X.; Chen, K.; Zheng, Y.; Zhou, Z.; Li, R.; Zhan, J.; Zhou,
  Y.; Li, L.; Yang, X.; Wu, L.; Yin, Z.; Huang, X.; and Qiu, X. 2023.
\newblock MOSS: Training Conversational Language Models from Synthetic Data.

\bibitem[{Touvron et~al.(2023)Touvron, Lavril, Izacard, Martinet, Lachaux,
  Lacroix, Rozière, Goyal, Hambro, Azhar, Rodriguez, Joulin, Grave, and
  Lample}]{touvron2023llama}
Touvron, H.; Lavril, T.; Izacard, G.; Martinet, X.; Lachaux, M.-A.; Lacroix,
  T.; Rozière, B.; Goyal, N.; Hambro, E.; Azhar, F.; Rodriguez, A.; Joulin,
  A.; Grave, E.; and Lample, G. 2023.
\newblock LLaMA: Open and Efficient Foundation Language Models.
\newblock arXiv:2302.13971.

\bibitem[{Wang et~al.(2018)Wang, Singh, Michael, Hill, Levy, and
  Bowman}]{wang-etal-2018-glue}
Wang, A.; Singh, A.; Michael, J.; Hill, F.; Levy, O.; and Bowman, S. 2018.
\newblock {GLUE}: A Multi-Task Benchmark and Analysis Platform for Natural
  Language Understanding.
\newblock In \emph{Proceedings of the 2018 {EMNLP} Workshop {B}lackbox{NLP}:
  Analyzing and Interpreting Neural Networks for {NLP}}, 353--355. Brussels,
  Belgium: Association for Computational Linguistics.

\bibitem[{Wang et~al.(2022)Wang, Zhang, Zhang, Yang, Gao, Wu, Dong, He, Zhuo,
  Yang, Huang, Li, Wu, Lu, Zhu, Chen, Han, Pan, Wang, Wang, Wu, Zeng, Chen,
  Gan, and Zhang}]{fengshenbang}
Wang, J.; Zhang, Y.; Zhang, L.; Yang, P.; Gao, X.; Wu, Z.; Dong, X.; He, J.;
  Zhuo, J.; Yang, Q.; Huang, Y.; Li, X.; Wu, Y.; Lu, J.; Zhu, X.; Chen, W.;
  Han, T.; Pan, K.; Wang, R.; Wang, H.; Wu, X.; Zeng, Z.; Chen, C.; Gan, R.;
  and Zhang, J. 2022.
\newblock Fengshenbang 1.0: Being the Foundation of Chinese Cognitive
  Intelligence.
\newblock \emph{CoRR}, abs/2209.02970.

\bibitem[{Williams, Nangia, and Bowman(2018)}]{Williams2018MNLI}
Williams, A.; Nangia, N.; and Bowman, S. 2018.
\newblock A Broad-Coverage Challenge Corpus for Sentence Understanding through
  Inference.
\newblock In \emph{Proceedings of the 2018 Conference of the North American
  Chapter of the Association for Computational Linguistics: Human Language
  Technologies, Volume 1 (Long Papers)}, 1112--1122. Association for
  Computational Linguistics.

\bibitem[{Xu et~al.(2020)Xu, Hu, Zhang, Li, Cao, Li, Xu, Sun, Yu, Yu, Tian,
  Dong, Liu, Shi, Cui, Li, Zeng, Wang, Xie, Li, Patterson, Tian, Zhang, Zhou,
  Liu, Zhao, Zhao, Yue, Zhang, Yang, Richardson, and Lan}]{xu-etal-2020-clue}
Xu, L.; Hu, H.; Zhang, X.; Li, L.; Cao, C.; Li, Y.; Xu, Y.; Sun, K.; Yu, D.;
  Yu, C.; Tian, Y.; Dong, Q.; Liu, W.; Shi, B.; Cui, Y.; Li, J.; Zeng, J.;
  Wang, R.; Xie, W.; Li, Y.; Patterson, Y.; Tian, Z.; Zhang, Y.; Zhou, H.; Liu,
  S.; Zhao, Z.; Zhao, Q.; Yue, C.; Zhang, X.; Yang, Z.; Richardson, K.; and
  Lan, Z. 2020.
\newblock {CLUE}: A {C}hinese Language Understanding Evaluation Benchmark.
\newblock In \emph{Proceedings of the 28th International Conference on
  Computational Linguistics}, 4762--4772. Barcelona, Spain (Online):
  International Committee on Computational Linguistics.

\bibitem[{Zeng et~al.(2022)Zeng, Liu, Du, Wang, Lai, Ding, Yang, Xu, Zheng, Xia
  et~al.}]{zeng2022glm}
Zeng, A.; Liu, X.; Du, Z.; Wang, Z.; Lai, H.; Ding, M.; Yang, Z.; Xu, Y.;
  Zheng, W.; Xia, X.; et~al. 2022.
\newblock Glm-130b: An open bilingual pre-trained model.
\newblock \emph{arXiv preprint arXiv:2210.02414}.

\bibitem[{Zeng(2023)}]{zeng2023mmcu}
Zeng, H. 2023.
\newblock Measuring Massive Multitask Chinese Understanding.
\newblock arXiv:2304.12986.

\bibitem[{Zhang et~al.(2022)Zhang, Roller, Goyal, Artetxe, Chen, Chen, Dewan,
  Diab, Li, Lin, Mihaylov, Ott, Shleifer, Shuster, Simig, Koura, Sridhar, Wang,
  and Zettlemoyer}]{zhang2022opt}
Zhang, S.; Roller, S.; Goyal, N.; Artetxe, M.; Chen, M.; Chen, S.; Dewan, C.;
  Diab, M.; Li, X.; Lin, X.~V.; Mihaylov, T.; Ott, M.; Shleifer, S.; Shuster,
  K.; Simig, D.; Koura, P.~S.; Sridhar, A.; Wang, T.; and Zettlemoyer, L. 2022.
\newblock OPT: Open Pre-trained Transformer Language Models.
\newblock arXiv:2205.01068.

\bibitem[{Zheng et~al.(2023)Zheng, Chiang, Sheng, Zhuang, Wu, Zhuang, Lin, Li,
  Li, Xing, Zhang, Gonzalez, and Stoica}]{zheng2023llm-judge}
Zheng, L.; Chiang, W.-L.; Sheng, Y.; Zhuang, S.; Wu, Z.; Zhuang, Y.; Lin, Z.;
  Li, Z.; Li, D.; Xing, E.~P.; Zhang, H.; Gonzalez, J.~E.; and Stoica, I. 2023.
\newblock Judging LLM-as-a-judge with MT-Bench and Chatbot Arena.
\newblock arXiv:2306.05685.

\bibitem[{Zhong et~al.(2023)Zhong, Cui, Guo, Liang, Lu, Wang, Saied, Chen, and
  Duan}]{Zhong2023AGIEval}
Zhong, W.; Cui, R.; Guo, Y.; Liang, Y.; Lu, S.; Wang, Y.; Saied, A. S.~S.;
  Chen, W.; and Duan, N. 2023.
\newblock AGIEval: A Human-Centric Benchmark for Evaluating Foundation Models.
\newblock \emph{ArXiv}, abs/2304.06364.

\bibitem[{Zhuang et~al.(2021)Zhuang, Wayne, Ya, and
  Jun}]{zhuang-etal-2021-roberta}
Zhuang, L.; Wayne, L.; Ya, S.; and Jun, Z. 2021.
\newblock A Robustly Optimized {BERT} Pre-training Approach with Post-training.
\newblock In \emph{Proceedings of the 20th Chinese National Conference on
  Computational Linguistics}, 1218--1227. Huhhot, China: Chinese Information
  Processing Society of China.

\end{thebibliography}


\begin{thebibliography}{9}
\bibliographystyle{apalike}

\bibitem{[1]} OpenAI. 2023. GPT-4 Technical Report. arXiv:2303.08774.  %\newline 

\bibitem{[2]} Cobbe, Karl and Kosaraju, Vineet and Bavarian, Mohammad and Chen, Mark and Jun, Heewoo and Kaiser, Lukasz and Plappert, Matthias and Tworek, Jerry and Hilton, Jacob and Nakano, Reiichiro and others,  2021. Training verifiers to solve math word problems.
arXiv:2110.14168. %\newline 

\bibitem{[3]} Hendrycks, Dan and Burns, Collin and Kadavath, Saurav and Arora, Akul and Basart, Steven and Tang, Eric and Song, Dawn and Steinhardt, Jacob,  2021. Measuring mathematical problem solving with the math dataset.
arXiv:2103.03874. %\newline 

\bibitem{[4]} Zheng, Lianmin and Chiang, Wei-Lin and Sheng, Ying and Zhuang, Siyuan and Wu, Zhanghao and Zhuang, Yonghao and Lin, Zi and Li, Zhuohan and Li, Dacheng and Xing, Eric and others,  2023. Judging LLM-as-a-judge with MT-Bench and Chatbot Arena.
arXiv:2306.05685. %\newline 

\bibitem{[5]} Li, Xuechen and Zhang, Tianyi and Dubois, Yann and Taori, Rohan and Gulrajani, Ishaan and Guestrin, Carlos and Liang, Percy and Hashimoto, Tatsunori B,  2023. Alpacaeval: An automatic evaluator of instruction-following models.
GitHub repository. %\newline 

\bibitem{[6]} Keisuke Sakaguchi and Ronan Le Bras and Chandra Bhagavatula and Yejin Choi,  2019. An Adversarial Winograd Schema Challenge at Scale.
arXiv:1907.10641. %\newline 


\bibitem{[7]} Peter Clark and Isaac Cowhey and Oren Etzioni and Tushar Khot and Ashish Sabharwal and Carissa Schoenick and Oyvind Tafjord,  2018. Think you have Solved Question Answering? Try ARC, the AI2 Reasoning Challenge.
arXiv:1803.05457. %\newline 


\end{thebibliography}
\end{document}